\documentclass[10pt,twocolumn,letterpaper]{article}

\usepackage{iccv}
\usepackage{times}
\usepackage{epsfig}
\usepackage{graphicx}
\usepackage{amsmath}
\usepackage{amssymb}
\usepackage{amsthm}
\usepackage{etoolbox}
\usepackage{tabulary}
\usepackage{xspace}
\usepackage{array,multirow}
\usepackage{boldline}

\usepackage[titlenumbered,ruled ]{algorithm2e}

\newcommand{\MyMapTemplatePrefixc}[4]{\expandafter#1\csname#3#4\endcsname{#2{#4}}} 
\forcsvlist{\MyMapTemplatePrefixc {\def} {\mathcal}{c}} {A,B,C,D,E,F,G,H,I,J,K,L,M,N,O,P,Q,R,S,T,U,V,W,X,Y,Z}  

\newcommand{\MyMapTemplatePrefixtb}[5]{\expandafter#1\csname#4#5\endcsname{#2{#3{#5}}}} 
\forcsvlist{\MyMapTemplatePrefixtb {\def} {\tilde}{\mathbf}{t}} {A,B,C,D,E,F,G,H,I,J,K,L,M,N,O,P,Q,R,S,T,U,V,W,X,Y,Z}  
\forcsvlist{\MyMapTemplatePrefixtb {\def} {\tilde}{\mathbf}{t}} {0,1,a,b,c,d,e,f,g,h,i,j,k,l,m,n,o,p,q,r,s,u,v,w,x,y,z}  

\newcommand{\MyMapTemplateNoPrefix}[3]{\expandafter#1\csname#3\endcsname{#2{#3}}}
\forcsvlist{\MyMapTemplateNoPrefix {\def} {\mathbf} } {0,1,a,b,c,d,e, f, g, h, i, j, k, l, m, n, o, p, q, r, u, v, w, x, y, z} 
\forcsvlist{\MyMapTemplateNoPrefix {\def} {\mathbf} } {A,B,C,D,E,F,G,H,I,J,K,L,M,N,O,P,Q,R,S,T,U,V,W,X,Y,Z}  

\usepackage[pagebackref=true,breaklinks=true,letterpaper=true,colorlinks,bookmarks=false]{hyperref}

\iccvfinalcopy 

\ificcvfinal\fi
\begin{document}

\title{Domain Adaptive Semantic Segmentation with Self-Supervised Depth Estimation}

\author{Qin Wang\textsuperscript{1} \quad Dengxin Dai\textsuperscript{1,2\thanks{The corresponding author}} \quad Lukas Hoyer\textsuperscript{1} \quad Luc Van Gool\textsuperscript{1,3} \quad Olga Fink\textsuperscript{1}\\
\textsuperscript{1}ETH Zurich, Switzerland \quad \textsuperscript{2}MPI for Informatics, Germany \quad \textsuperscript{3}KU Lueven, Belgium  \\
{\tt\small \{qwang,lhoyer,ofink\}@ethz.ch \{dai,vangool\}@vision.ee.ethz.ch}

}

\maketitle

\begin{abstract}
Domain adaptation for semantic segmentation aims to improve the model performance in the presence of a distribution shift between source and target domain. Leveraging the supervision from auxiliary tasks~(such as depth estimation) has the potential to heal this shift because many visual tasks are closely related to each other. However, such a supervision is not always available. In this work, we leverage the guidance from self-supervised depth estimation, which is available on both domains, to bridge the domain gap. On the one hand, we propose to explicitly learn the task feature correlation to strengthen the target semantic predictions with the help of target depth estimation. On the other hand, we use the depth prediction discrepancy from source and target depth decoders to approximate the pixel-wise adaptation difficulty. The adaptation difficulty, inferred from depth, is then used to refine the target semantic segmentation pseudo-labels. The proposed method can be easily implemented into existing segmentation frameworks. We demonstrate the effectiveness of our approach on the benchmark tasks SYNTHIA-to-Cityscapes and GTA-to-Cityscapes, on which we achieve the new state-of-the-art performance of $55.0\%$ and $56.6\%$, respectively. Our code is available at \url{https://qin.ee/corda}.

\end{abstract}

\section{Introduction}
The task of semantic segmentation requires models to assign pixel-level category labels to given scenes. While deep learning models have achieved good performance on benchmark datasets with the help of a large amount of high quality annotated training data~\cite{chen2017deeplab, yuan2019object}, they still face the real-world challenge of the domain shift between training and test data because of the variance in illumination, appearance, viewpoints, backgrounds, etc. Unsupervised domain adaptation~(UDA) can potentially heal this domain gap by aligning the domain distributions~\cite{vu2018advent}, or recursively refining the target pseudo-labels~\cite{zou2018domain}.

\begin{figure}[ht!]
	\centering
	\includegraphics[width=1.02\columnwidth ]{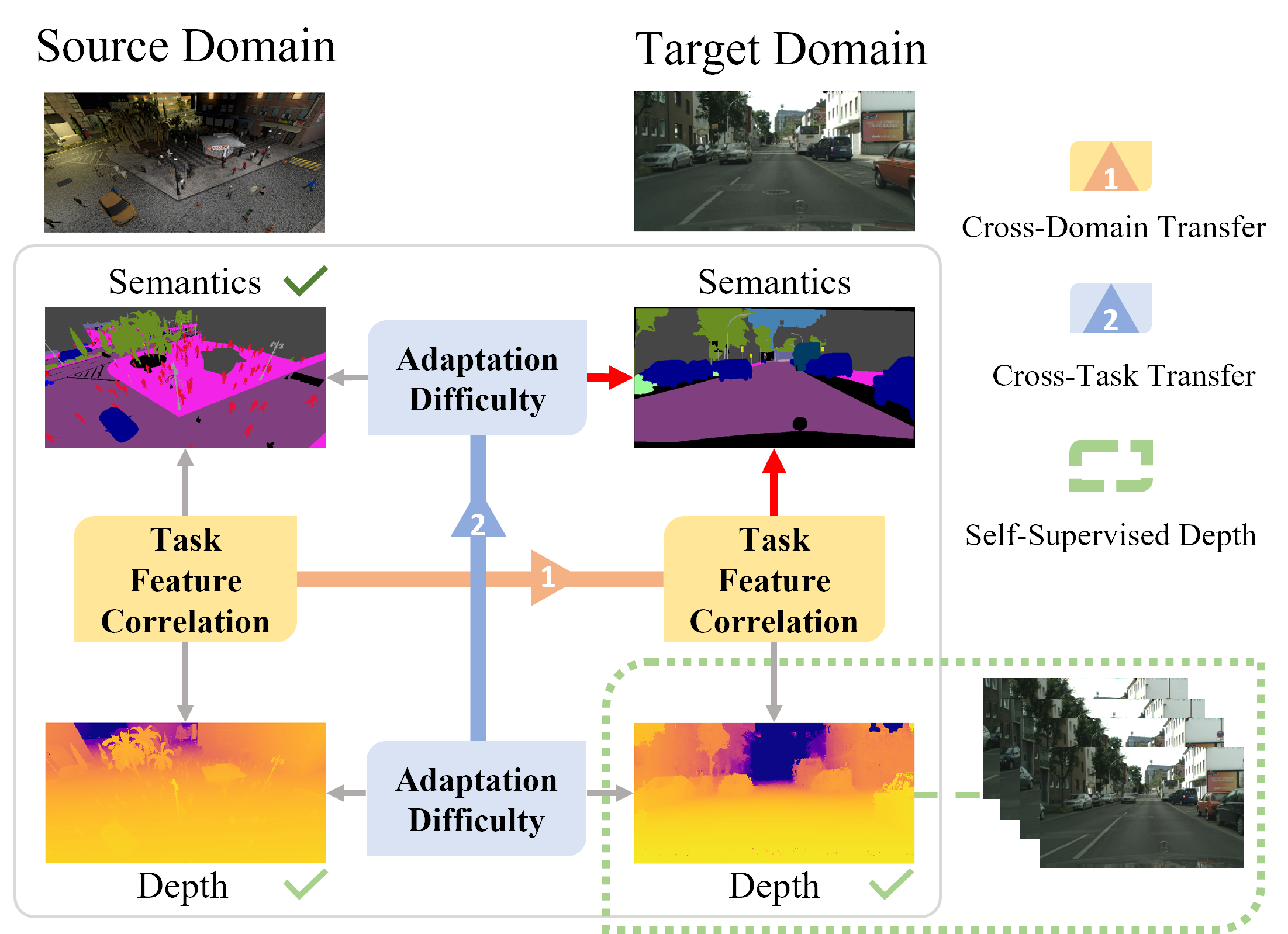}
	\caption{We propose to use self-supervised depth estimation~(green) to improve semantic segmentation performance under the unsupervised domain adaptation setup. We explicitly learn the \textit{task feature correlation}~(orange) between semantics and depth and use it to improve the target semantics. We use the \textit{adaptation difficulty}~(blue) approximated by depth prediction discrepancy of the target image from two domain-specific depth decoders to refine our target semantic pseudo-label. The proposed correlation-aware domain adaptation method can largely improve the segmentation performance in the target domain.}
	\label{view}
\end{figure}

In recent years, motivated by the success of multi-task learning~\cite{zamir2018taskonomy, xu2018pad},  auxiliary tasks~(such as depth estimation) have been increasingly used to help the adaptation. As auxiliary tasks are often coupled with the semantics, they have been proved to be beneficial for the main segmentation task~\cite{lee2018spigan}. Existing works~\cite{vu2019dada, chen2019learning} typically utilize the easy-to-access depth information from a synthetic source domain to train an auxiliary depth network but do not take target depth into account because of its inaccessibility. Inspired by recent progress on self-supervised depth estimation, where depth can be trained from stereo pairs~\cite{garg2016unsupervised, godard2017unsupervised} or video sequences~\cite{zhou2017unsupervised}, we propose to make use of self-supervised depth estimates for the domains (the source domain and/or target domain) on which ground-truth depth is not available.

The additional self-supervised depth estimation can facilitate us to explicitly learn the correlation between tasks to improve the final semantic segmentation performance. The learning of the correlation is motivated by the fact that the correlation between tasks is more invariant across domains than the individual modalities. As mentioned by previous works~\cite{chen2019learning}, sky is always faraway, roads and sideways are always flat. These domain-robust correlations between semantics and depth have the potential to largely improve the target semantic segmentation performance in the presence of a domain shift.

To this end, we propose to exploit such a correlation in two ways. On the one hand, we propose to explicitly learn the \textit{task feature correlation} between depth and semantics. This is achieved by using domain-shared multi-modal distillation modules to model the interaction and complementarity between semantics and depth features. The correlation learned from the source domain can be shared and transferred to the target domain to improve target segmentation performance. On the other hand, we make use of the correlation to refine the target semantic pseudo-labels. We approximate the \textit{adaptation difficulty} by calculating the discrepancy between the predictions of the domain-specific depth decoders. As depth and semantics are coupled, we make the assumption that the estimated adaptation difficulty can be transferred from depth to semantics. We propose to use this relation to guide the semantic segmentation pseudo-label refinement on the target domain. Combining the two ways of correlation exploitation leads to our proposed Correlation-Aware Domain Adaptation~(CorDA) approach. We illustrate the two ways to utilize the correlation in  Figure~\ref{view}.

It is also worth mentioning that our strategies can be implemented easily. The self-supervised depth estimation can be learned from easy-to-access image sequences or stereo images and the proposed correlation learning module can be readily incorporated into existing UDA networks for semantic segmentation. We demonstrate the effectiveness of our proposed approach on the benchmark tasks SYNTHIA-to-Cityscapes and GTA-to-Cityscapes, on which we achieve new state-of-the-art segmentation performance.

Our contributions are summarized as follows:
\begin{itemize}
	\item We propose a novel UDA framework which effectively utilizes self-supervised depth estimation available on both domains to improve semantic segmentation. 
	\item Specifically, we explicitly learn the correlation between modalities and share it across domains. Furthermore, we refine the semantic pseudo-labels by using the adaptation difficulty approximated by depth prediction discrepancy.
	\item Despite of the simplicity, our proposed approach achieves new state-of-the-art segmentation performance on the benchmark tasks SYNTHIA-to-Cityscapes and GTA-to-Cityscapes.
\end{itemize}

\section{Related Work}

\textbf{Unsupervised domain adaptation} 
Unsupervised domain adaptation~(UDA)~\cite{pan2011domain, patel2015visual} aims to improve the target model performance in the presence of a domain shift between the labeled source and unlabeled target domain. Many UDA methods have been proposed  to alleviate the domain shift. One common motivation is to align the source and target distribution~\cite{ganin2014unsupervised}.  This can be achieved in several different ways. AdaptSegNet~\cite{Tsai_adaptseg_2018} and Advent~\cite{vu2018advent} alleviates the domain shift by adversarially aligning the distributions in the output space or feature space. Another popular direction is to align the input pixels of source and target images via generative adversarial networks~\cite{hoffman2018cycada} or Fourier transforms~\cite{yang2020fda}. In recent years, especially in the field of UDA for semantic segmentation, pseudo-label refinement under a self-training frameworks has achieved competitive results. By iteratively using gradually-improving target pseudo-labels to train the network, the performance on the target domain can be further improved. Following this motivation, CBST~\cite{zou2018domain} improved the self-training performance by using class-specific thresholds. PyCDA~\cite{Lian_2019_ICCV} found that including pseudo-labels in different scales can further improve model performance. \cite{zheng2020rectifying} used the uncertainty of semantic predictions to refine the pseudo-labels. Using prototypes~\cite{zhang2021prototypical} to refine pseudo-labels has also shown promising results. Recently, DACS~\cite{tranheden2020dacs} demonstrated strong results by combining self-training with ClassMix~\cite{olsson2021classmix}, which mixes source and target images during the training. 

\textbf{Use of geometric information in semantic segmentation} 
Additional geometric information has been recently increasingly used to help learning the semantics~\cite{Ramirez_2019_ICCV} because geometric and semantic information are highly correlated. In the UDA setup, there are several works which pioneered this direction. SPIGAN~\cite{lee2018spigan} translates source images into the style of targets to reduce the domain gap. An auxiliary depth regression task is used in SPIGAN to regularize the generator, and better capture the semantics for the translated image. DADA~\cite{vu2019dada} uses an auxiliary depth prediction branch to predict the depth for both domains. The predictions are later fused together with semantic predictions and fed into the domain discriminator. GIO-Ada~\cite{chen2019learning} makes use of the depth information in both input-space translation and output-level adaptation, where a discriminator is applied on the concatenation of depth and semantic predictions. Existing works often use the additional depth information from the synthetic data in the source domain. The supervision from target geometric information is largely unexplored.

\textbf{Multi-task distillation}
Our work is also closely related to multi-task learning~(MTL)~\cite{mtl:survey}, where multiple tasks are predicted by a single network. Modern multi-task learning methods~\cite{xu2018pad, zhang2019pattern, vandenhende2020mti} aim at distilling the information from different tasks. This is often achieved by using a shared backbone network and task-specific heads. Initial task predictions are first made to learn task-specific intermediate features. These task-specific feature representations are then combined via a multi-modal distillation unit, before performing the final task predictions. Most multi-task learning works focus on the fully-supervised case where there exist no domain shift and the ground truths for all tasks are directly provided. We focus on the UDA setup where target ground truth is not provided for both main and auxiliary tasks. MTL under such a setup is understudied. Motivated by the success of these methods we modify and generalize the PAD-Net~\cite{xu2018pad} to capture the correlation between modalities across domains in order to facilitate the efficient joint learning of semantics and depth in the UDA setup. The idea of the multi-modality learning was also explored in other related areas such as object detection~\cite{liang2018deep, kong2018recurrent, ouyang2020dynamic}.

\textbf{Self-supervised learning}
Our work is also related to self-supervised learning in a broad sense. Self-supervised learning has recently achieved strong performance in learning meaningful representations in various vision tasks~\cite{he2020momentum, chen2020simple}. In the UDA for classification context, self-supervised learning has been shown to be able to improve generalization ability in the target domain by learning to predict auxiliary tasks~\cite{xu2019self, saito2020universal, sun2019unsupervised}. However, The auxiliary tasks used by these works are relatively arbitrary~(such as rotation prediction) and do not exploit the correlation between the main task and auxiliary task. In this work, we exploit the possibility of using depth estimation to improve the semantic segmentation performance under the UDA framework. In contrast to works combining (semi-)supervised semantic segmentation with self-supervised depth estimation~\cite{jiang2018self, hoyer2020three}, we explicitly deal with the challenge of the domain shift.

\section{Methodology}
In the UDA setup, we are given labeled data from the source domain and unlabeled training samples from the target domain. As annotations for synthetic data are comparably easy to generate, labeled synthetic data is often used as source $S$ and unlabeled target data is treated as target $T$. Formally, in the source domain, we have $\cD_S=\{(\x^S_{1}, \y^S_{1}, \mathbf{d}^S_1), \ldots, (\x^S_{n}, \y^S_{n}, \mathbf{d}^S_n)\}$ as the set of labeled training data, where $\x^S_i$ is the $i$-th sample, $\y^S_i$ is the corresponding label for semantic segmentation, $\mathbf{d}^S_i$ is the label for an optional auxiliary task (such as depth estimation), and $n$ is the total number of labeled source samples. The optional auxiliary task is not used in the classic UDA training setup. Similarly, target real training data can be represented as $\cD_{T}=\{(\x^{T}_1, \mathbf{d}^T_1), \ldots, (\x^T_m\, \mathbf{d}^T_m)\}$ where $\x^T_i$ is the $i$-th unlabeled training sample, $\mathbf{d}^T_i$ is the label for an optional auxiliary task, and $m$ is the number of unlabeled samples. The task of UDA for semantic segmentation is to train a model which performs well on test images $D_{test}=\{\x^{test}_1, \ldots, \x^{test}_t\}$ from the target domain $T$. We consider depth estimation as the auxiliary task.

Precise depth information is often not provided in the real-world target dataset. Existing works therefore often only use the source depth information from the virtual environment. Unfortunately, this limits the possibility of learning the comprehensive correlation between modalities and domains. To overcome this limitation, in this work, we propose to use self-supervised depth estimates as pseudo ground truth on the target domain $\mathbf{d}^T_i$. The use of self-supervised depth enables us to exploit the correlation between modalities to further improve the UDA performance as shown in Figure~\ref{view}.
First, we learn the domain-robust \textit{task feature correlation} between semantics and depth features on the source domain and transfer it with the target domain as described in Section~\ref{sec:corrarchitecture}. In our implementation, in order to avoid a two-stage training, we used a continuous transfer by having a shared module during the learning process. Second, we approximate the \textit{adaptation difficulty} by calculating the discrepancy between the predictions of source and target depth decoder. As depth and semantics are naturally coupled, we use the adaptation difficulty to refine the semantic pseudo labels as described in Section~\ref{sec:refine}.

\begin{figure*}
	\centering
	\includegraphics[width=0.95\textwidth ]{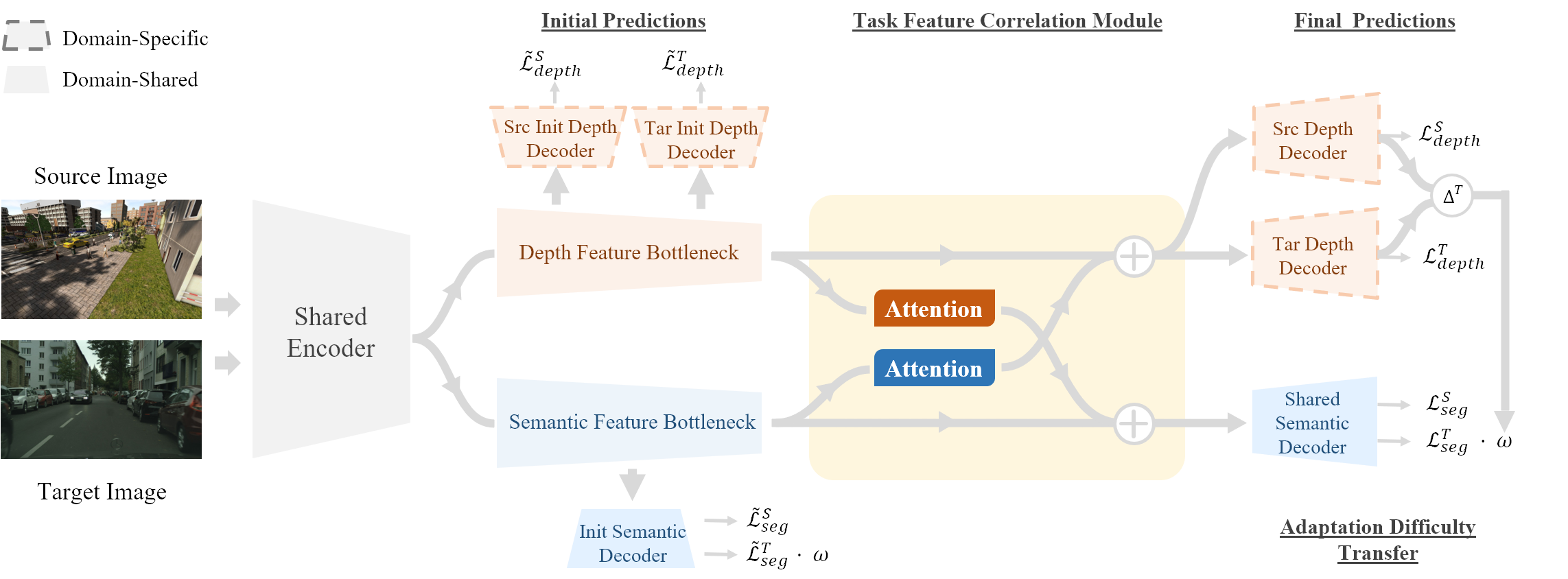}
	\caption{The network architecture of our proposed Correlation-Aware Domain Adaptation~(CorDA), in which we combine the proposed \textit{task feature correlation} module and the pseudo-label refinement based on \textit{adaptation difficulty transfer}. The semantic and depth features are processed by the domain-shared feature correlation module to explicitly learn the domain-robust correlation between them and provide complementary information for the other modality. In addition, as shown in the right-most side of the figure, during the training process, the semantic pseudo-labels are re-weighted based on the adaptation difficulty approximated by the depth prediction discrepancy.}
	\label{damain}
	\vspace{-5mm}
\end{figure*}

\subsection{Self-Supervised Depth Estimation}
The self-supervised depth estimation can be trained from stereo pairs~\cite{garg2016unsupervised, godard2017unsupervised} or video sequences~\cite{zhou2017unsupervised}. Both are relatively easy to obtain and, therefore, often already part of real-world datasets. By using off-the-shelf solutions such as Semi-Global Matching~\cite{hirschmuller2005accurate} and MonoDepth2~\cite{monodepth2}, pseudo depth information can be easily generated. The generated depth is used as the fixed pseudo depth ground truth for the training of our proposed model. A detailed explanation on the generation process is provided in Section~\ref{sec:data} and more extensively in the supplementary. If depth information is unavailable in the source domain, such as for GTA5~\cite{richter2016playing}, the same generation procedure can be applied  as well. These additional depth estimates can now facilitate the learning of correlation between semantics and depth in both domains.

\subsection{Correlation-Aware Architecture}
\label{sec:corrarchitecture}
In order to exploit the domain-robust correlation between the depth and the semantic information, we adapt recent developments of multi-task learning~\cite{xu2018pad, vandenhende2020mti} to our correlation-aware UDA framework. Figure~\ref{damain} depicts the framework of the proposed approach. Both domains share a common convolutional backbone network to encode images into deep features. This can be achieved by any modern deep CNN model. Then, domain-specific depth heads and a shared semantic prediction head are used to generate intermediate multi-modal predictions. In the next step, a domain-shared task feature correlation module is used to explicitly learn the correlation between depth and semantics and incorporate the complementary information from the other task to strengthen final segmentation predictions. 

\textbf{Domain-specific intermediate predictions}
Intermediate predictions are first generated to enable the later learning of the correlation between semantic and depth information. By applying convolution bottlenecks on the backbone features, we acquire semantic features and depth features of 256 channels. Semantic and depth prediction heads are applied to provide the intermediate predictions. We use two separate depth heads for source and target domains as depth supervision from both domains are available with the help of self-supervised depth estimation. Since there is no strong supervision available for target semantic predictions, we share the semantic heads for both domains. Predictions are re-scaled to the input resolution by bilinear interpolation. Following~\cite{vu2019dada}, we use the reverse Huber loss for depth:
\begin{equation}
\vspace{-0.1cm}
\text{berHu}(e_z)=\begin{cases}
\left|e_z\right|,& \text{if } \left|e_z\right|\leq c,\\
\frac{e_z^2+c^2}{2c}              & \text{otherwise},
\end{cases}
\vspace{-0.1cm}
\end{equation}
where $c$ is the threshold which is set to $\frac{1}{5}$ of the maximum depth difference. We use the cross entropy loss for the semantic loss calculation. This leads us to the following loss components for the intermediate prediction losses:

\begin{align}
\vspace{-0.1cm}
\mathcal{\tilde{L}}_{seg}^S( {\x^S},  {\y^S})&=-\sum_{h=1}^H\sum_{w=1}^W  { \y}^S \log \hat{\y}_{init}^S,\\
\mathcal{\tilde{L}}_{seg}^T( {\x^T},  {\y^T})&=-\sum_{h=1}^H\sum_{w=1}^W w \tilde{\y}^T \log \hat{\y}_{init}^T,
\label{eq:w}\\
\mathcal{\tilde{L}}_{depth}^S( {\x^S},  {\mathbf{d}^S})&= \sum_{h=1}^H\sum_{w=1}^W \text{berHu}(\hat{\mathbf{d}}^S_{ {init}} - {\mathbf{d}^S}),\\
\mathcal{\tilde{L}}_{depth}^T( {\x^T},  {\mathbf{d}^T})&= \sum_{h=1}^H\sum_{w=1}^W \text{berHu}(\hat{\mathbf{d}}^T_{ {init}} - {\mathbf{d}^T}),
\vspace{-0.1cm}
\end{align}
where $\hat{\y}_{init}$ are the semantic intermediate predictions. $\tilde{\y}^T$ is the one-hot semantic pseudo-label for target domain. Both $\hat{\y}_{init}^S$ and $\hat{\y}_{init}^T$ are the intermediate semantic predictions from the same shared semantic decoder.  $\hat{\mathbf{d}}^S$ and $\hat{\mathbf{d}}^T$ are the  intermediate depth predictions from the separate source and target depth decoders. $w$ is a pixel-wise pseudo-label weight which we will introduce in Section~\ref{sec:refine}. Following~\cite{chen2019learning,vu2019dada}, inverse depth is adopted for the depth learning losses. For all our experiments, the ground-truth depth is either from the simulator or from pre-calculated depth estimations. The tilde in $\mathcal{\tilde{L}}$ indicates that this is the loss function for the intermediate predictions.

\textbf{Shared task feature correlation module}
The semantic and depth features from the last step are then fed into a domain-shared task feature correlation module to learn the correlation between semantics and depth. This is achieved by incorporating two spatial attentions, which capture the mutual relationship between depth and semantics. The design of the feature correlation module is largely inspired by works in the field of multi-task learning~\cite{vandenhende2020mti, xu2018pad}, where similar attention modules were used to help the joint learning of multiple tasks. Existing works extract the correlation from multiples scales and different modalities.  We build our module based on PAD-Net~\cite{xu2018pad} because of its simplicity and effectiveness. Specifically, given the semantic features $F_{seg}$ and the depth features $F_{depth}$, the distilled features $F_{seg}^o, F_{depth}^o,$ are calculated by:
\begin{align}
\vspace{-0.1cm}
F_{seg}^o  &= F_{seg} + 	(W_d^1 \otimes F_{depth})	\odot \sigma(W_{d}^2 \otimes F_{depth})\\
F_{depth}^o &= F_{depth} + 	(W_s^1 \otimes F_{seg})	\odot \sigma(W_{s}^2 \otimes F_{seg}),
\vspace{-0.1cm}
\end{align}
where $\otimes$ denotes the convolution operation and $\odot$ denotes the element-wise multiplication. $\sigma$ is the sigmoid function for the normalization of the attention map. $W$ denotes the learnable  weights for the convolution. We notice that this self-attention variant performs better in our experiments. 

The benefits of the task feature correlation module are twofold. On the one hand, the attention captures the complementary information from the other modality and ignores the irrelevant information. Thus, we explicitly learn the correlation between the two modalities. On the other hand, by designing to share the attentions from the source domain to the target domain, we aim to learn a more robust and more generalizable correlation.

\textbf{Domain-specific final decoders}
Given the distilled semantic features $F_{seg}^o$ and the distilled depth features $F_{depth}^o$, we can now provide the final predictions for the entire network. Similar to the intermediate predictions, we use a shared semantic decoder for both domains to perform final predictions using $F_{seg}^o$ as input. The depth decoders of source and target domain remain independent. The overall loss function for the entire network, thus, results in:

\begin{align}
\vspace{-0.06cm}
\begin{split}
\mathcal{L} &= \mathcal{\tilde{L}}_{seg}^S +  \mathcal{\tilde{L}}_{seg}^T + \alpha^{S}\mathcal{\tilde{L}}_{depth}^S  + \alpha^{T} \mathcal{\tilde{L}}_{depth}^T \\ &+ \mathcal{L}_{seg}^S +  \mathcal{L}_{seg}^T + \alpha^{S}\mathcal{L}_{depth}^S  + \alpha^{T}\mathcal{L}_{depth}^T ,
\end{split}
\vspace{-0.06cm}
\label{eq:main}
\end{align}
where $\alpha^{S}$ and $\alpha^{T}$ are the hyperparameters for the depth loss. The loss functions for the final predictions have the same formulations as their intermediate counterparts.

\textbf{Summary of the architecture}
The architecture of the proposed framework, thus, contains domain-specific depth decoders and a shared task feature correlation module to explicitly learn the correlation between the depth and semantics. The final semantic predictions $\hat{y}^T$ for the target images can then be generated by the semantic decoder. 

\subsection{Pseudo-Label Refinement with Adaptation Difficulty}
\label{sec:refine}
As target semantic supervision is unavailable in the UDA setup, it is common for self-training approaches~\cite{zou2019confidence, Lian_2019_ICCV} to use target semantic predictions $\hat{y}^T$ as semantic pseudo-labels $\tilde{y}^T$ for the training. However, pseudo-labels can be noisy and over-confident~\cite{zou2019confidence}, thus it is important to filter out unreliable ones. Existing works refine pseudo-labels by exploiting prediction uncertainty~\cite{zheng2020rectifying} and class-wise confidence~\cite{zou2019confidence}. Our method is complementary to them. We leverage the availability of self-supervised depth and task correlations to refine the semantic pseudo-labels. 

With domain-specific depth decoders, we can approximate the difficulty of domain adaptation by calculating the discrepancy between the predictions of source and target depth decoders on the target image. As depth and semantics are naturally coupled, we assume that the estimated adaptation difficulty can be transferred from depth to semantics. We exploit this relation to refine the semantic pseudo labels.

Specifically, given a target image input $\x^T$, we calculate the final depth predictions of the target image using both the source depth decoder $f^S$ and the target depth decoder $f^T$. We compare the pixel-wise prediction discrepancy between the depth estimated by both the source and target decoder. The discrepancy is then used as an indicator for the pixel-wise adaptation difficulty. We hypothesize that the adaptation difficulty can be transferred from depth to semantics because of the coupled relationship of semantic and depth. Pixels where the depth prediction discrepancy is high indicate a larger domain gap for this region, thus, should be assigned a lower weight for the semantic pseudo-labels. The following equation is used to assign weights for the semantic pseudo-labels on the target domain: 

\begin{align}
\vspace{-0.06cm}
\begin{split}
\Delta &= \text{abs}( f^S(\x^T) - f^T(\x^T) ) \\
w &= \text{relu}(1 - \frac{\Delta}{ \mathbf{d}^T}),
\end{split}
\vspace{-0.06cm}
\end{align}
where $\mathbf{d}^T$ is the pseudo ground truth of the target depth and the pixel-wise weight $w$ is applied on target semantic pseudo labels in  $\mathcal{\tilde{L}}_{seg}^T $ and $\mathcal{L}_{seg}^T$, as shown in Equation~\ref{eq:w}.

The prediction difference is normalized by the pseudo ground truth $d^T$ in order to make the prediction difference more comparable across pixels with different distances with respect to the camera. The pixel-wise weight $w$ is designed to be in the range 0 to 1. If the source and target depth decoders give identical predictions for a pixel in the target image, this indicates that the domain gap in this region is very small, and the predicted semantic pseudo-label is likely to be correct. Thus, we assign 1 to the semantic pseudo-label for this pixel. If the depth prediction difference is large, then the domain gap is large, thus, it is hard to predict the semantics correctly. In this case, the weight $w$ becomes closer to 0, and the semantic pseudo-label for this region has little contribution to the semantic training loss. 

\subsection{Summary}
\label{sec:summary}
Combining the proposed correlation-aware architecture with the \textit{task feature correlation} transfer and the pseudo-label refinement with \textit{adaptation difficulty} leads to our Correlation-Aware Domain Adaptation~(CorDA) framework. As shown in Figure~\ref{damain}, we use a correlation-aware architecture, which incorporates a shared feature correlation module and domain-specific depth decoders. During the entire training process, the semantic pseudo-labels are re-weighted using the pixel-wise domain gap indicator introduced in our depth-guided difficulty refinement. 

The proposed method can be readily integrated into any UDA framework for semantic segmentation. To show that our method is complementary to existing frameworks, we use DACS~\cite{tranheden2020dacs} as our base framework as it offers a simple but strong baseline. DACS  mixes source and target images and uses a fixed threshold to filter the pseudo-labels. 

\section{Experiments}
\label{sec:data}
 We evaluate our proposed approach on the benchmark tasks SYNTHIA-to-Cityscapes and GTA-to-Cityscapes.
\paragraph{Cityscapes} The Cityscapes dataset~\cite{Cordts2016Cityscapes} is a real-world dataset containing driving scenarios of European cities. It contains fine semantic segmentations with 19 classes and consists of 2,975 training images as well as 500 validation images. Following the experimental protocol used by~\cite{chen2019learning}, the original images which have a fixed spatial resolution of 2048 × 1024 pixels are down-sized to 1024 × 512.  We use the publicly-available stereo depth estimation from~\cite{sakaridis2018model}. These depth estimations were originally generated using the Semi-Global Matching~\cite{hirschmuller2005accurate} with stereoscopic inpainting~\cite{wang2008stereoscopic}. In the ablation study, we also evaluate the possibility of using self-supervised monocular depth estimation as pseudo ground truth for Cityscapes. It is provided by a Monodepth2~\cite{monodepth2} model trained on the Cityscapes training image sequences. We use the Cityscapes training set without labels as target domain for the adaptation and report our results on the validation set. We always report the Intersection Over Union~(IoU) for per class performance as well as the mean Intersection over Union~(mIoU) over all classes.

\paragraph{SYNTHIA} The SYNTHIA dataset~\cite{ros2016SYNTHIA} is a synthetic dataset of road scenes collected from a virtual environment. Following the setup used by~\cite{vu2019dada, chen2019learning}, we adopt the SYNTHIA-RAND-CITYSCAPES split using Cityscapes-style annotations~(16 overlapping classes). The dataset consists of 9,400 synthetic images. We use the simulated depth provided by the dataset as our source depth supervision. 

\paragraph{GTA5} The GTA5 dataset~\cite{richter2016playing} is generated from a game environment. It contains 24,966 images which are labeled using Cityscapes-style annotation~(19 classes). We use Monodepth2~\cite{monodepth2} to generate the depth information for the GTA5 dataset. The monodepth2 model is trained solely on the image sequences from GTA5 dataset. We will release our monocular depth estimation datasets.

\begin{table}[h]
\small
\centering
\caption{Ablation study of different components in our proposed framework on the SYNTHIA-to-Cityscapes adaptation task. Stereo depth estimation is used for the target data. mIoU* denotes performance over 13 classes excluding wall, fence, and pole as it is also widely used in the literature.}\label{tab:ablation}

\vspace{2mm}
\scalebox{1.0}{
\tabcolsep7pt
\begin{tabular}{l|c|c|c|c|c}
\hline
  Method & \rotatebox[origin=c]{90}{Depth}& \rotatebox[origin=c]{90}{\parbox{1.4cm}{\centering Feature\\ Corr.}}& \rotatebox[origin=c]{90}{\parbox{1.4cm}{\centering Difficulty\\ Refine.}}& mIoU* & mIoU\\ \hline
Baseline~\cite{tranheden2020dacs} &  & & & 54.8 & 48.3 \\\hline
SimpleAux&\checkmark & & & 55.9&49.6\\
CorDA~(F)&\checkmark&\checkmark&&	62.4&	54.2

 \\
CorDA~(FD) &\checkmark&\checkmark&\checkmark& 	\textbf{62.8}&	\textbf{55.0}
  \\\hline

\hline
\end{tabular}}
\vspace{-3mm}
\end{table}

\paragraph{Implementation details} 
For our correlation-aware architecture, we adopt ResNet-101~\cite{he2016deep}
as the shared encoder and DeepLabv2~\cite{chen2017deeplab} as task decoder. The semantic and depth feature bottlenecks are residual blocks with two 3x3 and four 1x1 convolution operations.  Our training procedure is based on DACS~\cite{tranheden2020dacs} and enhanced by our pseudo-label refinement with adaptation difficulty.
Following~\cite{tranheden2020dacs}, batch size is set as 2. The learning rate starts from $2.5\times 10^{-4}$ and  follows a polynomial decay with exponent of 0.9. Images from the source domain are scaled to $1280\times 760$. The resolution of $1024\times 512$ is used for the target domain as input for training. Random crops of size $512 \times 512$ are used as an additional augmentation. We set the weights for source depth loss to $\alpha^S = 0.01$ and target depth loss weight to $\alpha^T = 0.001$. For the GTA-to-Cityscapes task, we apply the initial semantic decoder after the ResNet features, and use the initial semantic prediction as pseudo labels for the first 10\% training iterations. This helps the model to learn in early stages. All models are trained for 250,000 iterations. We report our performance at the end of the training.

\begin{table*}[h!]
\small
\centering
\caption{Semantic segmentation results for the SYNTHIA-to-Cityscapes adaptation task. mIoU* denotes performance over 13 classes excluding those marked with *. }\label{tab:syn2city}

\scalebox{0.8}{
\tabcolsep6pt
\begin{tabular}{l|l|cccccccccccccccc|c|c}
\hline
  Method & \rotatebox[origin=c]{90}{Depth}& \rotatebox[origin=c]{90}{road} & \rotatebox[origin=c]{90}{s.walk} & \rotatebox[origin=c]{90}{build.} & \rotatebox[origin=c]{90}{wall*} & \rotatebox[origin=c]{90}{fence*} & \rotatebox[origin=c]{90}{pole*} & \rotatebox[origin=c]{90}{light} & \rotatebox[origin=c]{90}{sign} & \rotatebox[origin=c]{90}{veget.} & \rotatebox[origin=c]{90}{sky} & \rotatebox[origin=c]{90}{person} & \rotatebox[origin=c]{90}{rider} & \rotatebox[origin=c]{90}{car} & \rotatebox[origin=c]{90}{bus} & \rotatebox[origin=c]{90}{moto.} & \rotatebox[origin=c]{90}{bike} & mIoU* & mIoU\\ \hline
  Source~\cite{tranheden2020dacs} & &36.3 & 14.6 & 68.8 & 9.2 & 0.2 & 24.4 & 5.6 & 9.1 & 69.0 & 79.4 & 52.5 & 11.3 & 49.8 & 9.5 & 11.0 & 20.7 & 33.7 & 29.5 \\
   OutputAdapt~\cite{Tsai_adaptseg_2018} && 84.3 & 42.7 & 77.5 & - & - & - & 4.7 & 7.0 & 77.9 & 82.5 & 54.3 & 21.0 & 72.3 & 32.2 & 18.9 & 32.3 & 46.7& -  \\
   ADVENT~\cite{vu2018advent} & &  85.6 & 42.2 & 79.7 & 8.7 & 0.4 & 25.9 & 5.4 & 8.1 & 80.4 & 84.1 & 57.9 & 23.8 & 73.3 & 36.4 & 14.2 & 33.0  & 48.0 & 41.2\\
   CBST \cite{zou2018domain}   & & 68.0 & 29.9 & 76.3 & 10.8 & 1.4 & 33.9 & 22.8 & 29.5 & 77.6 & 78.3 & 60.6 & 28.3 & 81.6 & 23.5 & 18.8 & 39.8 & 48.9 & 42.6 \\ 
R-MRNet \cite{zheng2020rectifying}   & & 87.6 & 41.9 & 83.1 & 14.7 & 1.7 & 36.2 & 31.3 & 19.9 & 81.6 & 80.6 & 63.0 & 21.8 & 86.2 & 40.7 & 23.6 & 53.1 & 54.9 & 47.9 \\ 

SIM \cite{wang2020differential} & & 83.0 & 44.0 & 80.3 & - & - & - & 17.1 & 15.8 & 80.5 & 81.8 & 59.9 & 33.1 & 70.2 & 37.3 & 28.5 & 45.8 & 52.1 & - \\ 

FDA \cite{yang2020fda} & & 79.3 & 35.0 & 73.2 & - & - & - & 19.9 & 24.0 & 61.7 & 82.6 & 61.4 & 31.1 & 83.9 & 40.8 &  \textbf{38.4} & 51.1 & 52.5 & - \\

Yang et al. \cite{yang2020labelddriven} & & 85.1 & 44.5 & 81.0 & - & - & - & 16.4 & 15.2 & 80.1 & 84.8 & 59.4 & 31.9 & 73.2 & 41.0 & 32.6 & 44.7 & 53.1 & - \\

IAST \cite{mei2020instance} & & 81.9 & 41.5 & 83.3 & 17.7 & 4.6 & 32.3 & 30.9 & 28.8 & 83.4 & 85.0 & 65.5 & 30.8 & \textbf{86.5} & 38.2 & 33.1 & 52.7 & 57.0 & 49.8  \\
DACS~\cite{tranheden2020dacs} & & 80.6 & 25.1 & 81.9 & 21.5 & 2.9 & 37.2 & 22.7 & 24.0 & 83.7 & 90.8 & 67.6 & 38.3 & 82.9 & 38.9 & 28.5 & 47.6 & 54.8 & 48.3 \\\hline
SPIGAN~\cite{lee2018spigan}&\checkmark&71.1&29.8&71.4&3.7&0.3&33.2&6.4&15.6&81.2&78.9&52.7&13.1&75.9&25.5&10.0&20.5&42.4&36.8\\
GIO-Ada~\cite{chen2019learning} &\checkmark&
78.3 & 29.2 & 76.9 & 11.4 &  0.3 & 26.5 & 10.8 & 17.2 & 81.7 & 81.9 & 45.8 & 15.4 & 68.0 & 15.9 &  7.5 & 30.4  & 43.0&  37.3 \\
DADA~\cite{vu2019dada} &\checkmark&89.2&44.8&81.4&6.8&0.3&26.2&8.6&11.1&81.8&84.0&54.7&19.3&79.7&40.7&14.0&38.8&49.8&42.6\\
CTRL~\cite{saha2021learning} &\checkmark&86.4&42.5&80.4&20.0&1.0&27.7&10.5&13.3&80.6&82.6&61.0&23.7&81.8&\textbf{42.9}&21.0&44.7&51.5&45.0\\\hline
CorDA~(mono) &\checkmark& 
90.2&	47.5&	\textbf{85.6}&	\textbf{24.5}&	3.0&	\textbf{38.2}&	\textbf{41.6}&	36.5&	\textbf{85.9}&	\textbf{91.7}&	\textbf{70.3}&	\textbf{42.4}&	86.0&	\textbf{42.9}&	34.7&	50.4&	62.0&	54.5\\
CorDA~(stereo) &\checkmark& 
\textbf{93.3}&	\textbf{61.6}&	85.3&	19.6&	\textbf{5.1}&	37.8&	36.6&	\textbf{42.8}&	84.9&	90.4&	69.7&	41.8&	85.6&	38.4&	32.6&	\textbf{53.9}&	\textbf{62.8}&	\textbf{55.0}\\\hline

\hline
\end{tabular}}
\end{table*}

\subsection{Results on SYNTHIA$\xrightarrow{}$Cityscapes}
We first evaluate the effectiveness of the proposed model on the SYNTHIA-to-Cityscapes task. We report the mIoU performance on the common 16 classes. 

\paragraph{Ablation study: individual modules}

The main contribution of our proposed framework is to utilize the self-supervised depth to effectively learn the shared correlation between tasks and domains. To validate our motivation, we conduct an ablation study on each of these components. We first include DACS as a strong baseline, which is already able to capture semantics relatively well without the help of geometric information. We then additionally use the self-supervised depth and add the depth prediction auxiliary task (without using the task feature correlation module) for both source and target depth to check whether a naive approach can provide improvement to the DACS baseline. Then, we evaluate our proposed domain-shared task feature correlation module to verify the contribution of explicitly learning the correlation between modalities. Finally, we add our pseudo-label refinement based on the adaptation difficulty. This final setup corresponds to our proposed framework. 

As shown in Table~\ref{tab:ablation}, directly using source and target depth information as auxiliary tasks~(denoted as SimpleAux in the Table) without making any modifications on the architecture and training process can already lead to a small improvement over the DACS baseline and gives us 49.6\% mIoU. This verifies the common belief that additional depth information can be helpful for learning semantics. However, the improvement is not significant, most likely because this naive way of simultaneously learning two tasks can not guarantee a good generalization ability for both tasks~\cite{kokkinos2017ubernet,xu2018pad}. By explicitly modelling the correlation between depth and semantics using the correlation-aware architecture with task feature correlation CorDA~(F), we can make better use of the depth information and significantly reduce the domain gap. This leads to an 4.6\% absolute improvement, yielding 54.2\% mIoU on the target domain. If we remove the correlation learning modules and keep the extra feature and semantic bottlenecks, the performance drops back to 51.7\% mIoU. This clearly demonstrates the importance of learning the correlation between the two modalities. In addition, by comparing the prediction discrepancy between source and target depth decoders, we integrate our pseudo-label refinement with adaptation difficulty module into the network, which leads to our final proposed framework CorDA~(FD). This gives us further 0.8\% of absolute performance improvement. From Table~\ref{tab:ablation}, we can observe that both the \textit{correlation-aware architecture with task feature correlation} and \textit{ pseudo-label refinement with adaptation difficulty} are beneficial for improving the semantic segmentation performance. The results clearly validate contributions of each of the proposed components.

\begin{table*}[ht]
\centering
\caption{Experiment results (mIoU in \%) on the  GTA5-to-Cityscapes task. Our method CorDA uses monocular depth estimation for GTA5 and stereo depth estimation for Cityscapes. }\label{tab:gta2city}
\scalebox{0.77}{
\tabcolsep5pt
\begin{tabulary}{\textwidth}{l|ccccccccccccccccccc|c}
\hline
Method  & \rotatebox[origin=c]{90}{road} & \rotatebox[origin=c]{90}{s.walk} & \rotatebox[origin=c]{90}{build.} & \rotatebox[origin=c]{90}{wall} & \rotatebox[origin=c]{90}{fence} & \rotatebox[origin=c]{90}{pole} & \rotatebox[origin=c]{90}{light} & \rotatebox[origin=c]{90}{sign} & \rotatebox[origin=c]{90}{veget.} & \rotatebox[origin=c]{90}{terrain} & \rotatebox[origin=c]{90}{sky} & \rotatebox[origin=c]{90}{person} & \rotatebox[origin=c]{90}{rider} & \rotatebox[origin=c]{90}{car} & \rotatebox[origin=c]{90}{truck} & \rotatebox[origin=c]{90}{bus}& \rotatebox[origin=c]{90}{train} & \rotatebox[origin=c]{90}{moto.} & \rotatebox[origin=c]{90}{bike} & mIoU\\ \hline
  Source only~\cite{Tsai_adaptseg_2018}  & 75.8 & 16.8 & 77.2 & 12.5 & 21.0 & 25.5 & 30.1 & 20.1 & 81.3 & 24.6 & 70.3 & 53.8 & 26.4 & 49.9 & 17.2 & 25.9 & 6.5 & 25.3 & 36.0 & 36.6 \\
  ROAD~\cite{Chen_2018_CVPR}  & 76.3 & 36.1 & 69.6 & 28.6 & 22.4 & 28.6 & 29.3 & 14.8 & 82.3 & 35.3 & 72.9 & 54.4 & 17.8 & 78.9 & 27.7 & 30.3 & 4.0 & 24.9 & 12.6 & 39.4\\
  OutputAdapt~\cite{Tsai_adaptseg_2018}  & 86.5 & 36.0 & 79.9 & 23.4 & 23.3 & 23.9 & 35.2 & 14.8 & 83.4 & 33.3 & 75.6 & 58.5 & 27.6 & 73.7 & 32.5 & 35.4 & 3.9 & 30.1 & 28.1 & 42.4 \\
  ADVENT~\cite{vu2018advent} & 87.6 & 21.4 & 82.0 &  34.8 & 26.2 & 28.5 & 35.6 & 23.0 & 84.5 & 35.1 & 76.2 & 58.6 & 30.7 & 84.8 & 34.2 & 43.4 & 0.4 & 28.4 & 35.3 & 44.8 \\
CBST \cite{zou2018domain}   & 91.8 & 53.5 & 80.5 & 32.7 & 21.0 & 34.0 & 28.9 & 20.4 & 83.9 & 34.2 & 80.9 & 53.1 & 24.0 & 82.7 & 30.3 & 35.9 & 16.0 & 25.9 & 42.8 & 45.9 \\ 
BDL \cite{li2019bidirectional}   & 91.0 & 44.7 & 84.2 & 34.6 & 27.6 & 30.2 & 36.0  & 36.0 & 85.0 & 43.6 & 83.0 & 58.6 & 31.6 & 83.3 & 35.3 & 49.7 & 3.3 & 28.8 & 35.6 & 48.5 \\ 
MRKLD-SP \cite{zou2019confidence} & 90.8 & 46.0 & 79.9 & 27.4 & 23.3 & \textbf{42.3} & 46.2 & 40.9 & 83.5 & 19.2 & 59.1 & 63.5 & 30.8 & 83.5 & 36.8 & 52.0 & 28.0 & 36.8 & 46.4 & 49.2  \\ 



Kim et al. \cite{DBLP:conf/cvpr/KimB20a} & 92.9 & 55.0 & 85.3 & 34.2 & 31.1 & 34.9 & 40.7 & 34.0 & 85.2 & 40.1 & 87.1 & 61.0 & 31.1 & 82.5 & 32.3 & 42.9 & 0.3 & 36.4 & 46.1 & 50.2 \\ 

CAG-UDA \cite{DBLP:conf/nips/ZhangZ0T19} & 90.4 & 51.6 & 83.8 & 34.2 & 27.8 & 38.4 & 25.3 & 48.4 & 85.4 & 38.2 & 78.1 & 58.6 & 34.6 & 84.7 & 21.9 & 42.7 & \textbf{41.1} & 29.3 & 37.2 & 50.2 \\ 

FDA \cite{yang2020fda} & 92.5 & 53.3 & 82.4 & 26.5 & 27.6 & 36.4 & 40.6 & 38.9 & 82.3 & 39.8 & 78.0 & 62.6 & 34.4 & 84.9 & 34.1 & 53.1 & 16.9 & 27.7 & 46.4 & 50.5 \\ 
PIT \cite{lv2020PIT} & 87.5 & 43.4 & 78.8 & 31.2 & 30.2 & 36.3 & 39.9 & 42.0 & 79.2 & 37.1 & 79.3 & 65.4 & \textbf{37.5} & 83.2 & 46.0 & 45.6 & 25.7 & 23.5 & 49.9 & 50.6 \\ 
IAST \cite{mei2020instance} & 93.8 & 57.8 & 85.1 & \textbf{39.5} & 26.7 & 26.2 & 43.1 & 34.7 & 84.9 & 32.9 & 88.0 & 62.6 & 29.0 & 87.3 & 39.2 & 49.6 & 23.2 & 34.7 & 39.6 & 51.5 \\
DACS~\cite{tranheden2020dacs} & 89.9 & 39.7 & \textbf{87.9} & 30.7 & 39.5 & 38.5 & 46.4 & \textbf{52.8} & \textbf{88.0} & 44.0 & 88.8 & \textbf{67.2} & 35.8 & 84.5 & 45.7 & 50.2 & 0.0  & 27.3 & 34.0 & 52.1 \\ \hline
CorDA & \textbf{94.7}&	\textbf{63.1}&	87.6&	30.7&	\textbf{40.6}&	40.2&	\textbf{47.8}&	51.6&	87.6&	\textbf{47.0}&	\textbf{89.7}&	66.7 &	35.9&\textbf{	90.2}&	\textbf{48.9}&	\textbf{57.5}&	0.0& \textbf{39.8} &	\textbf{56.0}&	 \textbf{56.6}
 \\ \hline

 \hline

\end{tabulary}}

\end{table*}

\paragraph{Ablation study: choice of pseudo depth ground truth}
As mentioned in earlier sections, the depth information used as pseudo ground truth  can come from a variety of sources, such as self-supervised monocular depth estimation or stereoscopic depth estimation. In this ablation study, we compare the impact of the choice of the source of depth information and  investigate the robustness of our proposed method against different types of depth estimation. We again use SYNTHIA-to-Cityscapes as our evaluation task. We change the pseudo depth ground truth of Cityscapes from the before stereoscopic estimation to monocular depth estimation from Monodepth2. As shown in Table~\ref{tab:syn2city}, the performance of our complete model CorDA is relatively similar with the two depth options. The use of monocular depth yields 54.5\% mIoU, while the stereoscopic depth yields 55.0\% mIoU. Model performance with monocular depth  is slightly lower because the stereo depth usually has  higher estimation quality. In both cases, the performance is very competitive and much stronger than the baseline. This indicates that the proposed method is relatively robust to the choice of pseudo depth ground truth and is able to capture the correlation between semantics and depth information, regardless of whether it is a monocular or stereo estimation. We would like to highlight that for both stereo and monocular depth estimations, only stereo pairs or image sequences from the same dataset are used to train and generate the pseudo depth estimation model. As no data from external datasets is used, and stereo pairs and image sequences are relatively easy to obtain, our proposal of using self-supervised depth have the potential to be effectively realized in real-world applications.

\begin{figure}[t!]
	\centering
	\includegraphics[width=1.0\columnwidth ]{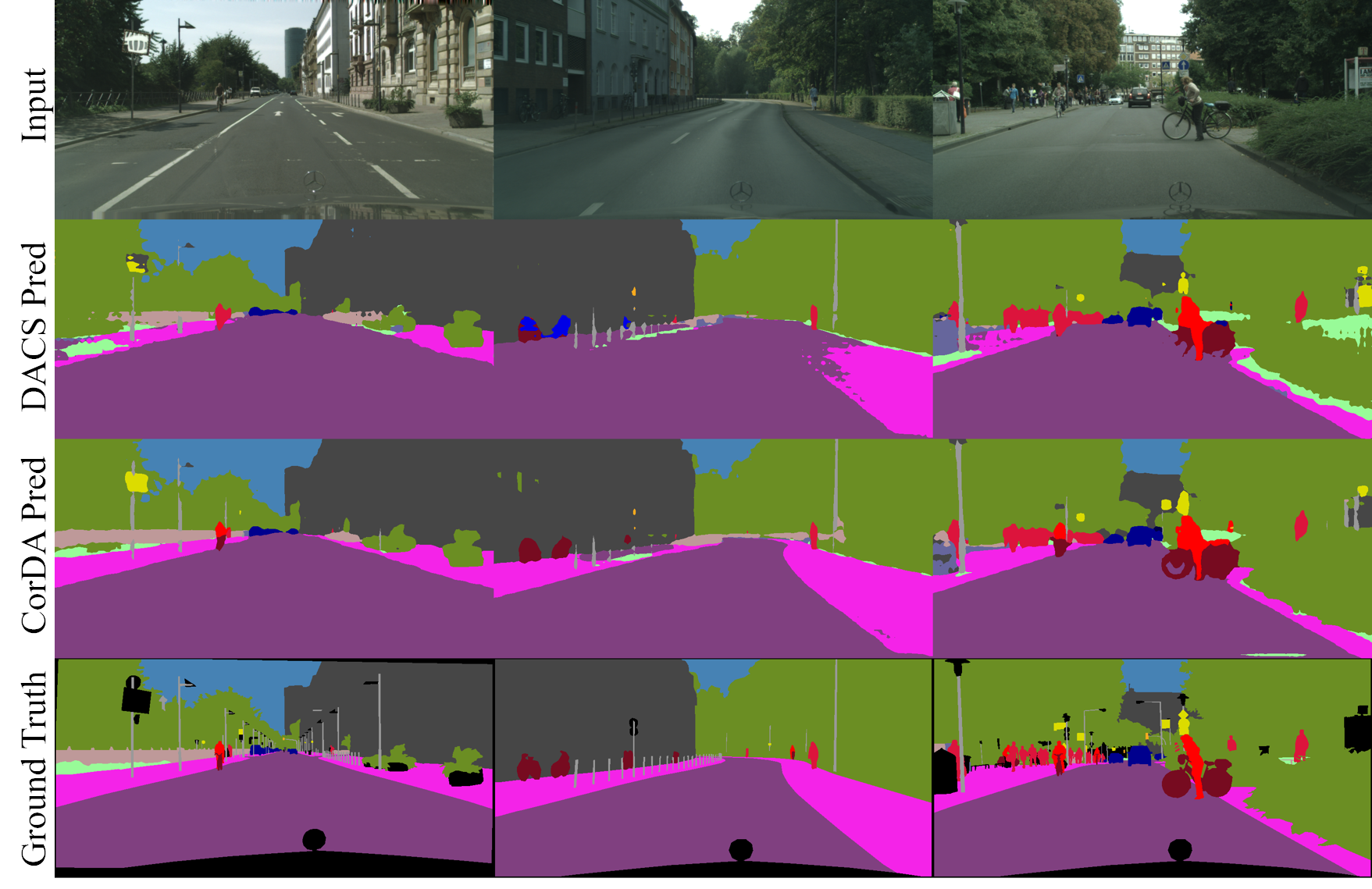}
	\caption{Semantic segmentation results on GTA-to-Cityscapes.}
	\label{quali}
	\vspace{-5mm}
\end{figure}
\paragraph{Comparison to the state-of-the-art approaches}
We compare the performance of our final proposed model to state-of-the-art methods on the SYNTHIA-to-Cityscapes unsupervised domain adaptation task in Table~\ref{tab:syn2city}. By exploiting the supervision from self-supervised depth estimation and learning the correlation between semantics and depth, the proposed method achieves 55.0\% mIoU~(stereo depth) on this task. This yields a large margin of 6.7\% absolute improvement compared to the previous state-of-the-art published work DACS~\cite{tranheden2020dacs}. We would like to highlight that, by using either monocular or stereo depth estimations, our proposed method steadily outperforms the other approaches by a large margin. This again shows the importance of learning the correlation between semantic and depth.

We additionally compare our method to four existing works which also utilizes available depth information during training. Unlike these works which use adversarial training to make use of the additional depth from source domain, we explicitly learn the correlation between modalities in both domains without any adversarial component. This makes the training more stable and exploits the correlation more effectively. As shown in the table, the proposed CorDA outperforms these methods by a large margin. Nevertheless, our method is complementary and can be potentially combined with these existing adversarial methods.

\subsection{Results on GTA$\xrightarrow{}$Cityscapes}
To further demonstrate the effectiveness of the proposed CorDA framework and the importance of explicitly learning the correlation between depth and semantics, we compare our method to a wide range of 12 competitive works on the GTA5-to-Cityscapes task. The experimental results are summarized in Table~\ref{tab:gta2city}. We use monocular depth estimation as pseudo depth ground truth for GTA5~(as no stereo pairs are available due to the limitation of the dataset) and stereo depth estimation for Cityscapes. The results demonstrate that our framework is robust to different sources of depth estimations and a competitive CorDA model can be successfully trained using different types of depth estimations for the two domains. Our method yields an absolute improvement of 4.5\% mIoU  over DACS, and achieves 56.6\% mIoU. This outperforms competing methods with a significant margin. As shown by sample predictions in Figure~\ref{quali}, the prediction quality is largely improved on easily confusable classes such as sidewalk and road.

\paragraph{Choice of Pre-trained Weight}
 To ensure a fair comparison with DACS, the same pretrained weights~(ImageNet+COCO) was used in previous experiments. An alternative is to use ImageNet-only pretrained weights. To evaluate the impact of the pretrained weight choice on CorDA, we reran the benchmark experiments with stereo Cityscapes depth estimation using the ImageNet-only weights. In this setup, CorDA achieves 54.6\%~(16 classes) and 56.4\% mIoU for SYNTHIA- and GTA-to-Cityscapes, respectively. This performance is very similar to the results with ImageNet+COCO  weight, and still outperforms competing methods with a large margin.

\section{Conclusions}
In this work, we introduced a new domain adaptation framework for semantic segmentation which effectively leverages the guidance from self-supervision of auxiliary task to bridge domain gaps. The proposed method explicitly learns the correlation between semantics and auxiliary tasks to better transfer this domain-shared knowledge to the target domain. To achieve this, a domain-shared task feature correlation module is used. We further made use of the adaptation difficulty, approximated by the prediction discrepancy from the domain depth decoders, to refine our segmentation predictions. By integrating our approach into an existing self-training framework, we achieved state-of-the-art performance on the two benchmark tasks SYNTHIA-to-Cityscapes and GTA-to-Cityscapes. The results verified our motivation and demonstrated the importance of capturing the correlation between modalities to improve semantic segmentation performance.

\noindent
\textbf{Acknowledgement} 
The contributions of Qin Wang and Olga Fink were funded by the Swiss National Science Foundation (SNSF) Grant no. PP00P2\_176878. This work is also funded by Toyota Motor Europe via the research project TRACE-Zurich.

{\small
\bibliographystyle{ieee_fullname}
\bibliography{egbib}

\begin{thebibliography}{10}\itemsep=-1pt

\bibitem{chen2017deeplab}
Liang-Chieh Chen, George Papandreou, Iasonas Kokkinos, Kevin Murphy, and Alan~L
  Yuille.
\newblock Deeplab: Semantic image segmentation with deep convolutional nets,
  atrous convolution, and fully connected crfs.
\newblock {\em IEEE Transactions on Pattern Analysis and Machine Intelligence},
  40(4):834--848, 2017.

\bibitem{chen2020simple}
Ting Chen, Simon Kornblith, Mohammad Norouzi, and Geoffrey Hinton.
\newblock A simple framework for contrastive learning of visual
  representations.
\newblock In {\em Proceedings of International Conference on Machine Learning
  (ICML)}, pages 1597--1607. PMLR, 2020.

\bibitem{chen2019learning}
Yuhua Chen, Wen Li, Xiaoran Chen, and Luc~Van Gool.
\newblock Learning semantic segmentation from synthetic data: A geometrically
  guided input-output adaptation approach.
\newblock In {\em Proceedings of the IEEE/CVF Conference on Computer Vision and
  Pattern Recognition (CVPR)}, pages 1841--1850, 2019.

\bibitem{Chen_2018_CVPR}
Yuhua Chen, Wen Li, and Luc Van~Gool.
\newblock Road: Reality oriented adaptation for semantic segmentation of urban
  scenes.
\newblock In {\em Proceedings of the IEEE/CVF Conference on Computer Vision and
  Pattern Recognition (CVPR)}, June 2018.

\bibitem{Cordts2016Cityscapes}
Marius Cordts, Mohamed Omran, Sebastian Ramos, Timo Rehfeld, Markus Enzweiler,
  Rodrigo Benenson, Uwe Franke, Stefan Roth, and Bernt Schiele.
\newblock The cityscapes dataset for semantic urban scene understanding.
\newblock In {\em Proceedings of the IEEE/CVF Conference on Computer Vision and
  Pattern Recognition (CVPR)}, 2016.

\bibitem{ganin2014unsupervised}
Yaroslav Ganin and Victor Lempitsky.
\newblock Unsupervised domain adaptation by backpropagation.
\newblock In {\em Proceedings of International Conference on Machine Learning
  (ICML)}, pages 1180--1189, 2015.

\bibitem{garg2016unsupervised}
Ravi Garg, Vijay~Kumar BG, Gustavo Carneiro, and Ian Reid.
\newblock Unsupervised cnn for single view depth estimation: Geometry to the
  rescue.
\newblock In {\em Proceedings of the European Conference on Computer Vision
  (ECCV)}, pages 740--756, 2016.

\bibitem{godard2017unsupervised}
Cl{\'e}ment Godard, Oisin Mac~Aodha, and Gabriel~J Brostow.
\newblock Unsupervised monocular depth estimation with left-right consistency.
\newblock In {\em Proceedings of the IEEE/CVF Conference on Computer Vision and
  Pattern Recognition (CVPR))}, pages 270--279, 2017.

\bibitem{monodepth2}
Cl{\'{e}}ment Godard, Oisin {Mac Aodha}, Michael Firman, and Gabriel~J.
  Brostow.
\newblock Digging into self-supervised monocular depth prediction.
\newblock In {\em Proceedings of the IEEE/CVF International Conference on
  Computer Vision (ICCV)}, October 2019.

\bibitem{he2020momentum}
Kaiming He, Haoqi Fan, Yuxin Wu, Saining Xie, and Ross Girshick.
\newblock Momentum contrast for unsupervised visual representation learning.
\newblock In {\em Proceedings of the IEEE/CVF Conference on Computer Vision and
  Pattern Recognition (CVPR)}, pages 9729--9738, 2020.

\bibitem{he2016deep}
Kaiming He, Xiangyu Zhang, Shaoqing Ren, and Jian Sun.
\newblock Deep residual learning for image recognition.
\newblock In {\em Proceedings of the IEEE/CVF Conference on Computer Vision and
  Pattern Recognition (CVPR)}, pages 770--778, 2016.

\bibitem{hirschmuller2005accurate}
Heiko Hirschmuller.
\newblock Accurate and efficient stereo processing by semi-global matching and
  mutual information.
\newblock In {\em Proceedings of the IEEE Conference on Computer Vision and
  Pattern Recognition (CVPR)}, volume~2, pages 807--814. IEEE, 2005.

\bibitem{hoffman2018cycada}
Judy Hoffman, Eric Tzeng, Taesung Park, Jun-Yan Zhu, Phillip Isola, Kate
  Saenko, Alexei Efros, and Trevor Darrell.
\newblock Cycada: Cycle-consistent adversarial domain adaptation.
\newblock In {\em Proceedings of International Conference on Machine Learning
  (ICML)}, pages 1989--1998. PMLR, 2018.

\bibitem{hoyer2020three}
Lukas Hoyer, Dengxin Dai, Yuhua Chen, Adrian K{\"o}ring, Suman Saha, and Luc
  Van~Gool.
\newblock Three ways to improve semantic segmentation with self-supervised
  depth estimation.
\newblock {\em arXiv preprint arXiv:2012.10782}, 2020.

\bibitem{jiang2018self}
Huaizu Jiang, Gustav Larsson, Michael Maire~Greg Shakhnarovich, and Erik
  Learned-Miller.
\newblock Self-supervised relative depth learning for urban scene
  understanding.
\newblock In {\em Proceedings of the European Conference on Computer Vision
  (ECCV)}, pages 19--35, 2018.

\bibitem{DBLP:conf/cvpr/KimB20a}
Myeongjin Kim and Hyeran Byun.
\newblock Learning texture invariant representation for domain adaptation of
  semantic segmentation.
\newblock In {\em Proceedings of the IEEE/CVF Conference on Computer Vision and
  Pattern Recognition (CVPR)}, pages 12972--12981. {IEEE}, 2020.

\bibitem{kokkinos2017ubernet}
Iasonas Kokkinos.
\newblock Ubernet: Training a universal convolutional neural network for low-,
  mid-, and high-level vision using diverse datasets and limited memory.
\newblock In {\em Proceedings of the IEEE/CVF Conference on Computer Vision and
  Pattern Recognition (CVPR)}, pages 6129--6138, 2017.

\bibitem{kong2018recurrent}
Shu Kong and Charless~C Fowlkes.
\newblock Recurrent scene parsing with perspective understanding in the loop.
\newblock In {\em Proceedings of the IEEE Conference on Computer Vision and
  Pattern Recognition}, pages 956--965, 2018.

\bibitem{lee2018spigan}
Kuan-Hui Lee, German Ros, Jie Li, and Adrien Gaidon.
\newblock Spigan: Privileged adversarial learning from simulation.
\newblock In {\em Proceedings of International Conference on Learning
  Representations (ICLR)}, 2018.

\bibitem{li2019bidirectional}
Yunsheng Li, Lu Yuan, and Nuno Vasconcelos.
\newblock Bidirectional learning for domain adaptation of semantic
  segmentation.
\newblock In {\em Proceedings of the IEEE/CVF Conference on Computer Vision and
  Pattern Recognition (CVPR)}, pages 6936--6945, 2019.

\bibitem{Lian_2019_ICCV}
Qing Lian, Fengmao Lv, Lixin Duan, and Boqing Gong.
\newblock Constructing self-motivated pyramid curriculums for cross-domain
  semantic segmentation: A non-adversarial approach.
\newblock In {\em Proceedings of the IEEE/CVF International Conference on
  Computer Vision (ICCV)}, October 2019.

\bibitem{liang2018deep}
Ming Liang, Bin Yang, Shenlong Wang, and Raquel Urtasun.
\newblock Deep continuous fusion for multi-sensor 3d object detection.
\newblock In {\em Proceedings of the European Conference on Computer Vision
  (ECCV)}, pages 641--656, 2018.

\bibitem{lv2020PIT}
Fengmao Lv, Tao Liang, Xiang Chen, and Guosheng Lin.
\newblock Cross-domain semantic segmentation via domain-invariant interactive
  relation transfer.
\newblock In {\em Proceedings of the IEEE/CVF Conference on Computer Vision and
  Pattern Recognition (CVPR)}, pages 4334--4343, 2020.

\bibitem{mei2020instance}
Ke Mei, Chuang Zhu, Jiaqi Zou, and Shanghang Zhang.
\newblock Instance adaptive self-training for unsupervised domain adaptation.
\newblock In {\em Proceedings of the European conference on computer vision
  (ECCV)}, 2020.

\bibitem{olsson2021classmix}
Viktor Olsson, Wilhelm Tranheden, Juliano Pinto, and Lennart Svensson.
\newblock Classmix: Segmentation-based data augmentation for semi-supervised
  learning.
\newblock In {\em Proceedings of the IEEE/CVF Winter Conference on Applications
  of Computer Vision (WACV)}, pages 1369--1378, 2021.

\bibitem{ouyang2020dynamic}
Erli Ouyang, Li Zhang, Mohan Chen, Anurag Arnab, and Yanwei Fu.
\newblock Dynamic depth fusion and transformation for monocular 3d object
  detection.
\newblock In {\em Proceedings of the Asian Conference on Computer Vision},
  2020.

\bibitem{pan2011domain}
Sinno~Jialin Pan, Ivor~W Tsang, James~T Kwok, and Qiang Yang.
\newblock Domain adaptation via transfer component analysis.
\newblock {\em IEEE Transactions on Neural Networks}, 22(2):199--210, 2011.

\bibitem{patel2015visual}
Vishal~M Patel, Raghuraman Gopalan, Ruonan Li, and Rama Chellappa.
\newblock Visual domain adaptation: A survey of recent advances.
\newblock {\em IEEE signal processing magazine}, 32(3):53--69, 2015.

\bibitem{Ramirez_2019_ICCV}
Pierluigi~Zama Ramirez, Alessio Tonioni, Samuele Salti, and Luigi~Di Stefano.
\newblock Learning across tasks and domains.
\newblock In {\em Proceedings of the IEEE/CVF International Conference on
  Computer Vision (ICCV)}, October 2019.

\bibitem{richter2016playing}
Stephan~R Richter, Vibhav Vineet, Stefan Roth, and Vladlen Koltun.
\newblock Playing for data: Ground truth from computer games.
\newblock In {\em Proceedings of the European Conference on Computer Vision
  (ECCV)}, pages 102--118. Springer, 2016.

\bibitem{ros2016SYNTHIA}
German Ros, Laura Sellart, Joanna Materzynska, David Vazquez, and Antonio~M
  Lopez.
\newblock The synthia dataset: A large collection of synthetic images for
  semantic segmentation of urban scenes.
\newblock In {\em Proceedings of the IEEE/CVF Conference on Computer Vision and
  Pattern Recognition (CVPR)}, pages 3234--3243, 2016.

\bibitem{saha2021learning}
Suman Saha, Anton Obukhov, Danda~Pani Paudel, Menelaos Kanakis, Yuhua Chen,
  Stamatios Georgoulis, and Luc Van~Gool.
\newblock Learning to relate depth and semantics for unsupervised domain
  adaptation.
\newblock In {\em Proceedings of the IEEE/CVF Conference on Computer Vision and
  Pattern Recognition}, pages 8197--8207, 2021.

\bibitem{saito2020universal}
Kuniaki Saito, Donghyun Kim, Stan Sclaroff, and Kate Saenko.
\newblock Universal domain adaptation through self supervision.
\newblock In {\em Advances in Neural Information Processing Systems},
  volume~33, pages 16282--16292, 2020.

\bibitem{sakaridis2018model}
Christos Sakaridis, Dengxin Dai, Simon Hecker, and Luc Van~Gool.
\newblock Model adaptation with synthetic and real data for semantic dense
  foggy scene understanding.
\newblock In {\em Proceedings of the European Conference on Computer Vision
  (ECCV)}, pages 687--704, 2018.

\bibitem{sun2019unsupervised}
Yu Sun, Eric Tzeng, Trevor Darrell, and Alexei~A Efros.
\newblock Unsupervised domain adaptation through self-supervision.
\newblock {\em arXiv preprint arXiv:1909.11825}, 2019.

\bibitem{tranheden2020dacs}
Wilhelm Tranheden, Viktor Olsson, Juliano Pinto, and Lennart Svensson.
\newblock Dacs: Domain adaptation via cross-domain mixed sampling.
\newblock In {\em Proceedings of the IEEE/CVF Winter Conference on Applications
  of Computer Vision (WACV)}, pages 1379--1389, 2020.

\bibitem{Tsai_adaptseg_2018}
Y.-H. Tsai, W.-C. Hung, S. Schulter, K. Sohn, M.-H. Yang, and M. Chandraker.
\newblock Learning to adapt structured output space for semantic segmentation.
\newblock In {\em Proceedings of the IEEE/CVF Conference on Computer Vision and
  Pattern Recognition (CVPR)}, 2018.

\bibitem{mtl:survey}
S. {Vandenhende}, S. {Georgoulis}, W. {Van Gansbeke}, M. {Proesmans}, D. {Dai},
  and L. {Van Gool}.
\newblock Multi-task learning for dense prediction tasks: A survey.
\newblock {\em IEEE Transactions on Pattern Analysis and Machine Intelligence},
  2021.

\bibitem{vandenhende2020mti}
Simon Vandenhende, Stamatios Georgoulis, and Luc Van~Gool.
\newblock Mti-net: Multi-scale task interaction networks for multi-task
  learning.
\newblock In {\em Proceedings of the European Conference on Computer Vision
  (ECCV)}, pages 527--543. Springer, 2020.

\bibitem{vu2018advent}
Tuan-Hung Vu, Himalaya Jain, Maxime Bucher, Mathieu Cord, and Patrick
  P{\'e}rez.
\newblock Advent: Adversarial entropy minimization for domain adaptation in
  semantic segmentation.
\newblock In {\em Proceedings of the IEEE/CVF Conference on Computer Vision and
  Pattern Recognition (CVPR)}, 2019.

\bibitem{vu2019dada}
Tuan-Hung Vu, Himalaya Jain, Maxime Bucher, Matthieu Cord, and Patrick
  P{\'e}rez.
\newblock Dada: Depth-aware domain adaptation in semantic segmentation.
\newblock In {\em Proceedings of the IEEE/CVF International Conference on
  Computer Vision (ICCV)}, pages 7364--7373, 2019.

\bibitem{wang2008stereoscopic}
Liang Wang, Hailin Jin, Ruigang Yang, and Minglun Gong.
\newblock Stereoscopic inpainting: Joint color and depth completion from stereo
  images.
\newblock In {\em Proceedings of the IEEE Conference on Computer Vision and
  Pattern Recognition (CVPR)}, pages 1--8. IEEE, 2008.

\bibitem{wang2020differential}
Zhonghao Wang, Mo Yu, Yunchao Wei, Rogerio Feris, Jinjun Xiong, Wen-mei Hwu,
  Thomas~S Huang, and Honghui Shi.
\newblock Differential treatment for stuff and things: A simple unsupervised
  domain adaptation method for semantic segmentation.
\newblock In {\em Proceedings of the IEEE/CVF Conference on Computer Vision and
  Pattern Recognition (CVPR)}, pages 12635--12644, 2020.

\bibitem{xu2018pad}
Dan Xu, Wanli Ouyang, Xiaogang Wang, and Nicu Sebe.
\newblock Pad-net: Multi-tasks guided prediction-and-distillation network for
  simultaneous depth estimation and scene parsing.
\newblock In {\em Proceedings of the IEEE/CVF Conference on Computer Vision and
  Pattern Recognition (CVPR)}, pages 675--684, 2018.

\bibitem{xu2019self}
Jiaolong Xu, Liang Xiao, and Antonio~M L{\'o}pez.
\newblock Self-supervised domain adaptation for computer vision tasks.
\newblock {\em IEEE Access}, 7:156694--156706, 2019.

\bibitem{yang2020labelddriven}
Jinyu Yang, Weizhi An, Sheng Wang, Xinliang Zhu, Chaochao Yan, and Junzhou
  Huang.
\newblock Label-driven reconstruction for domain adaptation in semantic
  segmentation.
\newblock {\em Proceedings of the European conference on computer vision
  (ECCV)}, abs/2003.04614, 2020.

\bibitem{yang2020fda}
Yanchao Yang and Stefano Soatto.
\newblock Fda: Fourier domain adaptation for semantic segmentation.
\newblock In {\em Proceedings of the IEEE/CVF Conference on Computer Vision and
  Pattern Recognition (CVPR)}, pages 4085--4095, 2020.

\bibitem{yuan2019object}
Yuhui Yuan, Xilin Chen, and Jingdong Wang.
\newblock Object-contextual representations for semantic segmentation.
\newblock {\em Proceedings of the European Conference on Computer Vision
  (ECCV)}, 2020.

\bibitem{zamir2018taskonomy}
Amir~R Zamir, Alexander Sax, William Shen, Leonidas~J Guibas, Jitendra Malik,
  and Silvio Savarese.
\newblock Taskonomy: Disentangling task transfer learning.
\newblock In {\em Proceedings of the IEEE/CVF Conference on Computer Vision and
  Pattern Recognition (CVPR)}, pages 3712--3722, 2018.

\bibitem{zhang2021prototypical}
Pan Zhang, Bo Zhang, Ting Zhang, Dong Chen, Yong Wang, and Fang Wen.
\newblock Prototypical pseudo label denoising and target structure learning for
  domain adaptive semantic segmentation.
\newblock {\em arXiv preprint arXiv:2101.10979}, 2021.

\bibitem{DBLP:conf/nips/ZhangZ0T19}
Qiming Zhang, Jing Zhang, Wei Liu, and Dacheng Tao.
\newblock Category anchor-guided unsupervised domain adaptation for semantic
  segmentation.
\newblock In {\em Advances in Neural Information Processing Systems}, pages
  433--443, 2019.

\bibitem{zhang2019pattern}
Zhenyu Zhang, Zhen Cui, Chunyan Xu, Yan Yan, Nicu Sebe, and Jian Yang.
\newblock Pattern-affinitive propagation across depth, surface normal and
  semantic segmentation.
\newblock In {\em Proceedings of the IEEE/CVF Conference on Computer Vision and
  Pattern Recognition (CVPR)}, pages 4106--4115, 2019.

\bibitem{zheng2020rectifying}
Zhedong Zheng and Yi Yang.
\newblock Rectifying pseudo label learning via uncertainty estimation for
  domain adaptive semantic segmentation.
\newblock {\em International Journal of Computer Vision}, pages 1--15, 2020.

\bibitem{zhou2017unsupervised}
Tinghui Zhou, Matthew Brown, Noah Snavely, and David~G Lowe.
\newblock Unsupervised learning of depth and ego-motion from video.
\newblock In {\em Proceedings of the IEEE/CVF Conference on Computer Vision and
  Pattern Recognition (CVPR)}, pages 1851--1858, 2017.

\bibitem{zou2018domain}
Yang Zou, Zhiding Yu, BVK~Vijaya Kumar, and Jinsong Wang.
\newblock Unsupervised domain adaptation for semantic segmentation via
  class-balanced self-training.
\newblock In {\em Proceedings of the European Conference on Computer Vision
  (ECCV)}, pages 289--305, 2018.

\bibitem{zou2019confidence}
Yang Zou, Zhiding Yu, Xiaofeng Liu, B.~V. K.~Vijaya Kumar, and Jinsong Wang.
\newblock Confidence regularized self-training.
\newblock In {\em Proceedings of the IEEE/CVF International Conference on
  Computer Vision (ICCV)}, pages 5981--5990, 2019.

\end{thebibliography}
}

\end{document}